\documentclass[runningheads]{llncs}
\usepackage[T1]{fontenc}
\usepackage{graphicx}
\usepackage{multirow}
\usepackage{subcaption}
\usepackage{hhline}
\usepackage{amsmath}
\usepackage{amsfonts}
\usepackage[hidelinks]{hyperref}
\usepackage{algorithm}
\usepackage{algpseudocode}

\usepackage{orcidlink}

\newcommand\blfootnote[1]{%
  \begingroup
  \renewcommand\thefootnote{}\footnote{#1}%
  \addtocounter{footnote}{-1}%
  \endgroup
}

\begin{document}
\title{Adversarial Manhole: Challenging Monocular Depth Estimation and Semantic Segmentation Models with Patch Attack}
\titlerunning{Adversarial Manhole: Challenging MDE and SS with Patch Attack}
%
\author{Naufal Suryanto\inst{1}\orcidlink{0000-0002-0396-5938} \and
Andro Aprila Adiputra \inst{1}\orcidlink{0009-0000-9467-2841} \and
Ahmada Yusril Kadiptya \inst{1}\orcidlink{0009-0008-8612-9812} \and
Yongsu Kim\inst{1,2}\orcidlink{0000-0001-6169-5537} \and
Howon Kim\inst{1,2}\orcidlink{0000-0001-8475-7294}
}

\authorrunning{N. Suryanto et al.}
%
\institute{Pusan National University, Busan, South Korea \and
SmartM2M, Busan, South Korea
}
%
\maketitle              
\begin{abstract}

\vspace{-5mm}
Monocular depth estimation (MDE) and semantic segmentation (SS) are crucial for the navigation and environmental interpretation of many autonomous driving systems. However, their vulnerability to practical adversarial attacks is a significant concern.
This paper presents a novel adversarial attack using practical patches that mimic manhole covers to deceive MDE and SS models. The goal is to cause these systems to misinterpret scenes, leading to false detections of near obstacles or non-passable objects. We use Depth Planar Mapping to precisely position these patches on road surfaces, enhancing the attack's effectiveness.
Our experiments show that these adversarial patches cause a 43\% relative error in MDE and achieve a 96\% attack success rate in SS. These patches create affected error regions over twice their size in MDE and approximately equal to their size in SS.
Our studies also confirm the patch's effectiveness in physical simulations, the adaptability of the patches across different target models, and the effectiveness of our proposed modules, highlighting their practical implications.
\blfootnote{This research was supported by the MSIT(Ministry of Science and ICT), Korea, under the Convergence security core talent training business(Pusan National University) support program(RS-2022-II221201) supervised by the IITP(Institute for Information \& Communications Technology Planning \& Evaluation).}

\end{abstract}

\section{Introduction}
Self-driving cars rely on complex algorithms and are equipped with sensing devices such as cameras to navigate intricate environments. The core of a camera-based self-driving system utilizes computer vision techniques and deep learning to perceive objects around the car and make decisions. These cars must be constantly aware of their surroundings to ensure a safe driving environment. 

Deep learning-based computer vision technologies used in self-driving cars include Monocular Depth Estimation (MDE) and Semantic Segmentation (SS). MDE calculates object distances by extracting depth information from camera images, which is crucial for determining the distance ahead of the car. However, the impact of compromised MDE on autonomous driving tasks remains largely unknown \cite{cheng_physical_2022}. If MDE provides inaccurate depth information, it can increase the risk of collisions during split-second decisions.

Semantic Segmentation (SS), on the other hand, helps identify traffic objects such as vehicles, pedestrians, and traffic signs, which is vital for safe driving \cite{SurveyUnsupervisedSS}. If SS fails to accurately recognize these objects, it can lead to dangerous driving decisions and result in severe accidents.

Ensuring the robustness of MDE and SS methods is crucial to prevent critical failures. Adversarial attacks, which deceive models into producing incorrect outputs, pose a significant threat \cite{PhysicalAdversarialAttackSurvey}. These attacks often use methods like adversarial patches to manipulate predictions \cite{yamanaka_adversarial_2020,cheng_physical_2022,guesmi_aparate_2023,guesmi_saam_2024,nesti_evaluating_2022}. However, the practical implementation of such attacks is often overlooked. This work addresses this gap by focusing on realistic patch-based attacks targeting MDE and SS models with patches that resemble road manholes. Manholes, with their diverse designs, are ideal attack vectors. By using manhole-like patches, we highlight the real-world applicability and implications of our approach.

Our contributions can be summarized as follows:
\begin{itemize}
\item We present the Adversarial Manhole, a practical adversarial attack targeting MDE and SS in autonomous driving scenarios. To our knowledge, this is the first approach to attack both models simultaneously.
\item We propose a Depth Planar Mapping technique that utilizes depth information to accurately position adversarial patches on the road surface, going beyond simple perspective transformations.
\item Our framework generates a stealthy adversarial patch designed as a manhole cover, which can be deployed anywhere on the road, exploiting vulnerabilities in autonomous driving systems.
\item We conduct an extensive evaluation to assess the robustness of adversarial manholes across various scenarios, including physical simulations, applicability to different models, and ablation studies to analyze the effectiveness of our proposed modules.
\end{itemize}

\section{Related Works}

\subsection{Monocular Depth Estimation (MDE)}
Monocular Depth Estimation (MDE) can be achieved through various methods. Unsupervised methods, addressing labeling costs, treat MDE as a reconstruction problem using dense correspondence between stereo images. Self-supervised learning enhances this by extracting ego-motion, improving accuracy through automated pixel masking and projection error calculation \cite{monodepth2}. Another method sharpens depth representation for thin and close objects by enhancing photometric cues \cite{depthhints}. Techniques addressing inconsistencies between viewpoints, pose estimation, and cost volume loss further improve depth estimation accuracy \cite{manydepth}.

\subsection{Semantic Segmentation (SS)}
Semantic Segmentation (SS) assigns semantic labels to each pixel in an image, categorizing them into classes such as vehicles, pedestrians, roads, and buildings.

Deep learning, particularly convolutional neural networks, has significantly advanced image segmentation. PSPNet \cite{zhao_pyramid_2017} uses a Pyramid Pooling Module to capture rich contextual information but struggles with real-time performance. ICNet \cite{ferrari_icnet_2018} achieves real-time speed (30 FPS) with 70\% accuracy using an image cascade network. DDRNet \cite{ddrnet} improves upon this with dual-resolution networks, achieving 77.4\% mIoU at 102 FPS by using a novel DAPPM module to extract and merge multi-scale context information.

\subsection{Patch-based Adversarial Attacks}
Research on physical adversarial attacks targeting MDE includes generating noticeable adversarial patches \cite{yamanaka_adversarial_2020}, printable stealthy patches with distance limitations \cite{cheng_physical_2022}, and adaptive patches that sacrifice stealthiness for effectiveness \cite{guesmi_aparate_2023}. Another approach introduced stealthy patches resilient to transformations but limited to MDE \cite{guesmi_saam_2024}. Our work aims to target both MDE and SS.

Adversarial attacks methods on SS, like the Fast Gradient Sign Method (FGSM) have been explored but is impractical in real-world scenarios as they modify the entire pixel space 
\cite{arnab_robustness_2018}. Other approaches optimized adversarial patches using cross-entropy loss on pre-trained models, showing effectiveness in digital simulations but reduced performance in real-world applications \cite{nesti_evaluating_2022}.

Tab. \ref{tab:related_works} compares existing adversarial patch methods targeting MDE and SS with the proposed Adversarial Manhole. The comparison focuses on stealth capabilities, victim tasks, and patch projection methods. Our approach supports stealth features, pioneering simultaneous attacks in MDE and SS, and introducing Depth Planar Mapping for optimized road patch projection.

\begin{table}
\vspace{-5mm}
\caption{Comparison of proposed and existing patch-based adversarial attacks.}
\label{tab:related_works}
\centering
\begin{tabular}{|l|l|l|l|l}  
\hline
Method                                             & Stealthy        & Victim Task & Patch Projection                            \\ 
\hhline{|====|}
Adversarial Patch \cite{yamanaka_adversarial_2020} & No              & MDE         & Perspective Transformation            \\ 
\hline
SemSegAdvPatch \cite{nesti_evaluating_2022} & No              & SS          & Perspective Transformation            \\ 
\hline
OAP \cite{cheng_physical_2022}                     & Yes             & MDE         & Spatial Transformation                \\ 
\hline
APARATE \cite{guesmi_aparate_2023}                 & No              & MDE         & Spatial Transformation                \\ 
\hline
SAAM \cite{guesmi_saam_2024}                       & Yes             & MDE         & Perspective Transformation            \\ 
\hline
Adversarial Manhole (Ours)                         & Yes             & MDE, SS     & Depth Planar Mapping                  \\
\hline
\end{tabular}
\end{table}

\vspace{-7mm}
\section{Methodology}

\subsection{Problem Definition}

The primary objective of our proposed method is to create an adversarial manhole that, when placed on the road, causes monocular depth estimation (MDE) and semantic segmentation (SS) systems to misinterpret the scene. This misinterpretation should lead the vehicle to falsely detect a near obstacle or non-passable object, potentially causing it to stop or choose an alternate path.

Consider a benign input image $x$. In a patch-based adversarial attack, the goal is to alter the image by introducing an adversarial patch $\theta$, creating an adversarial example $x'$. To ensure realism, the patch is applied to the target surface using a patch mapping function $P$. The adversarial example with the patch is formulated as:
\begin{equation}
    x' = P(x, \theta, t_o, t_s) = x \odot (1 - m(t_o, t_s)) + M(\theta, t_o, t_s) \odot m(t_o, t_s),
\label{eq1:adv_patch_formulation}
\end{equation}
where $M$ maps the patch to the desired texture offset $t_o$ and texture scale $t_s$, $m$ provides a mask indicating where the patch is applied, and $\odot$ denotes element-wise multiplication.

For the MDE model, $D(\cdot)$, the input image $x$ is used to estimate the depth $d$. To manipulate the depth prediction towards a target depth $d_{\text{target}}$ using an adversarial patch $\theta$, we define a loss function $L_d$. The optimization problem is:
\begin{equation}
    \arg\min_{\theta} L_d\left(D(P(x, \theta, t_o, t_s)), d_{\text{target}}\right).
\label{eq2:adv_patch_mde_formulation}
\end{equation}

For the SS model, $S(\cdot)$, the input image $x$ is used to output the label $y$ for each pixel. We design a loss function $L_{ua}$ for an untargeted attack to predict an incorrect label $y' \neq y$, and $L_{ta}$ for a targeted attack to predict a desired label $y_{\text{target}}$. The optimization problem is:
\begin{equation}
    \arg\min_{\theta}
    \begin{cases}
        L_{ua}\left(S(P(x, \theta, t_o, t_s)), y\right), & \text{if untargeted} \\
        L_{ta}\left(S(P(x, \theta, t_o, t_s)), y_{\text{target}}\right). & \text{if targeted}
    \end{cases}
\label{eq3:adv_patch_ss_formulation}
\end{equation}

To simultaneously attack both $D(\cdot)$ and $S(\cdot)$, we combine the losses and apply Expectation over Transformation (EOT) \cite{athalye2018synthesizing} to enhance the robustness of the patch $\theta$ over different transformations $T$ and background images $X$. The combined optimization problem is:
\begin{align}
    \arg\min_{\theta} \mathbb{E}_{t\in T,x\in X} \bigg[ & \alpha \cdot L_d\left(D(P(x, \theta, t_o, t_s)), d_{\text{target}}\right) \nonumber \\
    & + \beta_{ua} \cdot L_{ua}\left(S(P(x, \theta, t_o, t_s)), y\right) \nonumber \\
    & + \beta_{ta} \cdot L_{ta}\left(S(P(x, \theta, t_o, t_s)), y_{\text{target}}\right) \bigg],
\label{eq4:loss_final}
\end{align}
where $\alpha$, $\beta_{ua}$, and $\beta_{ta}$ are hyperparameters controlling the influence of each loss.

\subsection{Adversarial Manhole}
\label{sec:adv_manhole}

\begin{figure*}[!t]
\begin{center}

\centerline{\includegraphics[width=\textwidth]{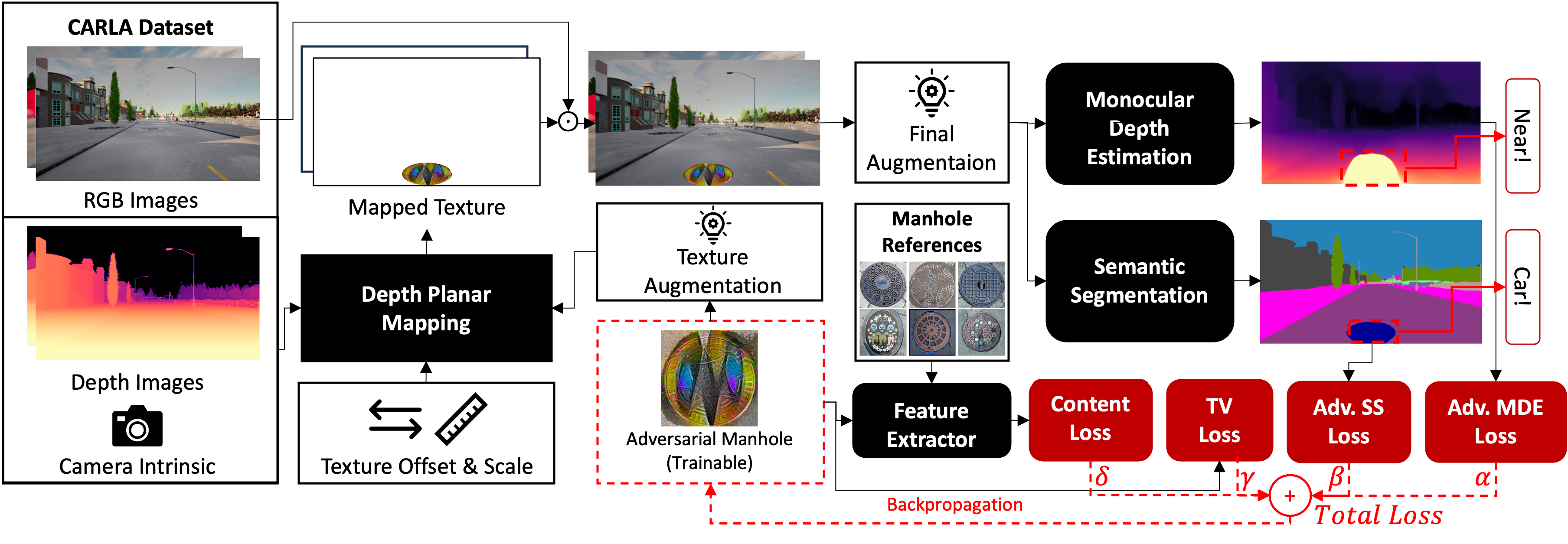}}

\caption{Our proposed framework for generating adversarial manhole.}
\label{fig:adv_manhole_framework}
\vspace{-1cm}
\end{center}
\end{figure*}

We propose the Adversarial Manhole Framework (Fig. \ref{fig:adv_manhole_framework}) to address these issues. The framework uses differentiable components and losses to optimize the adversarial manhole $\theta$ through gradient-based methods.



\vspace{-3mm}
\subsubsection{Depth Planar Mapping.} 
We propose Depth Planar Mapping to accurately map adversarial textures onto target surfaces, considering texture offset and scale as defined in Eq. \ref{eq1:adv_patch_formulation}. Inspired by \cite{Suryanto_2023_ICCV}, our approach uses depth information and camera intrinsics from the corresponding RGB image to calculate local surface coordinates and map the texture accordingly. Unlike their tri-planar mapping, we use single-planar (z-plane) mapping due to the upward normal of road surfaces. This method achieves more accurate texture mapping than typical perspective transformations in patch-based attacks. The detailed is described in Alg. \ref{alg1:depth_planar_mapping}.

\vspace{-3mm}
\subsubsection{Texture and Final Augmentation.} 
To enhance the robustness of the adversarial manhole, we apply digital augmentations as in the EOT method \cite{athalye2018synthesizing}. These include random brightness and contrast adjustments for various lighting conditions, and random flips and rotations for texture robustness. We formalize our Texture Augmentation ($A_t$) and Final Augmentation ($A_f$) as follows:
\begin{equation}
    \theta_t = A_t(\theta, t_T),
\label{eq:texture_augmentation}
\end{equation}
\vspace{-6.5mm}
\begin{equation}
    x'_t = A_f(x', t_F),
\label{eq:final_augmentation}
\end{equation}
where $t_T$ represents texture augmentation variables, and $t_F$ represents final augmentation variables.

\vspace{-3mm}
\subsubsection{Adversarial MDE Loss.} 
One goal of our adversarial manhole is to deceive the MDE model into detecting a near object in front of the vehicle. The MDE model typically outputs a disparity map, which is the inverse of depth: higher disparity values correspond to closer objects. We design the Adv. MDE Loss ($L_d$) to minimize Eq. \ref{eq2:adv_patch_mde_formulation} by enforcing zero distance predictions in the manhole area ($m$). The loss function is:
\begin{equation}
    L_{d}(x', m) = -\log(D_{disp}(x')) \odot m.
\label{eq:adv_patch_mde_loss}
\end{equation}
This maximizes the disparity in the manhole area, making the perceived distance close to zero.

\begin{figure}[t]
\vspace{-8mm}
\begin{algorithm}[H]
\caption{Depth Planar Mapping (DPM) Algorithm}
\label{alg1:depth_planar_mapping}
\begin{algorithmic}[1]
\State \textbf{Input:} Depth image $d$, Camera intrinsics $(c_x, c_y, f_x, f_y)$, Texture $\theta$, Texture scales $t_s$, Texture offsets $t_o$, Texture resolution $t_{res}$
\State \textbf{Output:} Mapped texture $M$ and Texture masks $m$

\State Calculate surface coordinates $SC_x, SC_y, SC_z$
\For {each pixel $(u, v)$ in depth image $d$}
    \State $SC_x = d(u, v) \cdot \frac{u - c_x}{f_x}$, $SC_y = d(u, v) \cdot \frac{v - c_y}{f_y}$, $SC_z = d(u, v)$
\EndFor

\State Form surface coordinates tensor (z-upward, x-forward): $SC = \{SC_z, SC_x, SC_y\}$

\State Adjust and normalize coordinates: $SC_{norm} = ((SC - t_o) \bmod t_s) / t_s$

\State Compute UV indices: $UV = \max(0, \min(\text{round}(SC_{norm} \cdot t_{res}), t_{res} - 1))$

\State Map texture to surface using z-plane: $M = \theta[UV_{x}, UV_{y}]$

\State Calculate bounds for masking: \\
$x_{min} = t_{o_x}$, $x_{max} = t_{o_x} + t_s$, $y_{min} = t_{o_y}$, $y_{max} = t_{o_y} + t_s$

\State Create texture mask: \\
$m = (SC_x > x_{min}) \& (SC_x < x_{max}) \& (SC_y > y_{min}) \& (SC_y < y_{max})$

\State \Return Mapped texture $M$ and Texture masks $m$
\end{algorithmic}
\end{algorithm} 
\vspace{-10mm}
\end{figure}

\vspace{-3mm}
\subsubsection{Adversarial SS Loss.} 
The adversarial manhole aims to trick the SS model into detecting an obstacle or non-passable object. For untargeted attacks, we use $L_{ua}$ to force predictions other than the road class, and for targeted attacks, $L_{ta}$ to achieve a desired class. The losses are:
\begin{equation}
    L_{ua}(x', m) = -\log(1 - S_{road}(x')) \odot m
\label{eq:adv_patch_ss_loss}
\end{equation}
\begin{equation}
    L_{ta}(x', m) = -\log(\max_{c \in C} S_{c}(x')) \odot m.
\label{eq:adv_patch_ss_ta_loss}
\end{equation}
These losses adjust the probabilities of the road class and target classes ($c \in C$) in the manhole area.

\vspace{-4mm}
\subsubsection{Total Variation Loss.} 
To ensure the adversarial manhole texture is smooth, we use Total Variation (TV) Loss \cite{mahendran2015understanding}, which minimizes abrupt color changes that are unrealistic in printed and captured images. The TV Loss ($L_{tv}$) is:
\begin{equation}
    L_{tv}(\theta) = \frac{1}{N} \sum_{i,j}^{N} |\theta_{i,j} - \theta_{i,j+1}| + |\theta_{i,j} - \theta_{i+1,j}|,
\label{eq:tv_loss}
\end{equation}
where $\theta_{i,j}$ is the pixel at index ($i,j$). This loss smooths the texture by minimizing differences between neighboring pixels.

\vspace{-3mm}
\subsubsection{Content Loss.} 
To make the adversarial manhole appear natural and more stealthy, we use Content Loss ($L_c$) \cite{cheng_physical_2022}, formulated as:
\begin{equation}
    L_{c}(\theta) = \min_{r \in R} |F_l(\theta) - F_l(r)|,
\label{eq:content_loss}
\end{equation}
where $F_l$ is the CNN feature extractor output at layer $l$, and $R$ is the set of reference manhole images. 
This loss minimizes the feature differences between the adversarial and reference images, ensuring visual similarity.

\begin{figure}[t]
\vspace{-8mm}
\begin{algorithm}[H]
\caption{Adversarial Manhole Generation}
\label{alg2:adv_manhole_generation}
\begin{algorithmic}[1]
\State \textbf{Input:} RGB Image $X$, Depth image $D$, Camera intrinsics $C$, Texture resolution $t_{res}$, Depth Planar Mapping $DPM$, Texture Augmentation $A_t$, Final Augmentation $A_f$, Random function $R$, Texture scales $t_S$, Texture offsets $t_O$, Texture transformations $t_T$, Final transformations $t_F$
\State \textbf{Output:} Adversarial Manhole $\theta$

\State Initialize $\theta$ with random values: $\theta = R(0, 1)$
\For {each iteration}
    \State Select a minibatch sample of data: $x \in X$, $d \in D$, $c \in C$
    \State Initialize random transformations: $t_s \in t_S$, $t_o \in t_O$, $t_t \in t_T$, $t_f \in t_F$
    \State Augment the manhole texture with $A_t$: $\theta_t = A_t(\theta, t_t)$ \Comment{Eq. \ref{eq:texture_augmentation}}
    \State Output the mapped texture $M$ and texture mask $m$ using $DPM$: \\ $M, m = DPM(d, c, \theta_t, t_s, t_o, t_{res})$ \Comment{Alg. \ref{alg1:depth_planar_mapping}}
    \State Combine the RGB image and mapped texture: $x' = x \odot (1 - m) + M \odot m$
    \State Augment the final image: $x'_t = A_f(x', t_f)$ \Comment{Eq. \ref{eq:final_augmentation}}
    \State Calculate losses: $L_d(x'_t, m)$, $L_{ua}(x'_t, m)$,  $L_{ta}(x'_t, m)$,  $L_{tv}(\theta)$, $L_c(\theta)$  \Comment{Eq.\ref{eq:adv_patch_mde_loss} - \ref{eq:content_loss}}
    \State Update $\theta$ to minimize the total loss $L_{total}$ via backpropagation \Comment{Eq. \ref{eq:final_loss}}
\EndFor
\State \Return Adversarial Manhole $\theta$
\end{algorithmic}
\end{algorithm}
\vspace{-10mm}
\end{figure}

\vspace{-3mm}
\subsubsection{Adversarial Manhole Generation.} 
We initialize $\theta$ with random values and generate an adversarial manhole using our framework. We iteratively update the texture to minimize the final loss ($L_{total}$):
\begin{equation}
    L_{total}(x', m, \theta) = \alpha \cdot L_d(x', m) + \beta_{ua} \cdot L_{ua}(x', m) + \beta_{ta} \cdot L_{ta}(x', m) + \gamma \cdot L_{tv}(\theta) + \delta \cdot L_c(\theta),
\label{eq:final_loss}
\end{equation}
where $\alpha$, $\beta_{ua}$, $\beta_{ta}$, $\gamma$, and $\delta$ are hyperparameters. The detailed is in Alg. \ref{alg2:adv_manhole_generation}.

\section{Experiments}

\subsection{Implementation Details}

\subsubsection{Datasets.} We generate autonomous driving scene datasets using CARLA \cite{carlaDosovitskiy17} including RGBs, depths, and camera intrinsics as described in Fig. \ref{fig:adv_manhole_framework}. We randomly selected positions from seven towns and captured the datasets resulting in 2,656 images. Each data has $1024 \times 780$ resolution from a 1-meter height camera. We split the dataset into train/val/test with a 60/20/20 ratio.

\vspace{-3mm}
\subsubsection{Frameworks.} We implemented our framework using PyTorch\footnote[1]{Code and datasets:  \url{https://github.com/naufalso/adversarial-manhole}.}, with MonoDepth2 \cite{monodepth2} as the target MDE model and DDRNet \cite{ddrnet} as the SS model. For adversarial manhole generation, we use the Adam optimizer with a learning rate of 0.01, batch size of 8, and 25 epochs. We set $T_{O_x} \in [0.0, 0.4]$ to randomly map the texture between 1.8 - 2.6m from the camera, and $T_{O_y} \in [-0.4, 0.4]$ to randomly map the texture between -0.8m and 0.8m from the center of the camera. For $t_F$ and $t_T$, we use 0.2 for random brightness and 0.1 for random contrast. The total loss hyperparameters are $\alpha=2.0$, $\beta_{ua}=0.5$, $\beta_{ta}=0.5$, $\gamma=1.0$, and $\delta=0.5$. For targeted attacks, we set the target label to non-walkable categories, such as buildings, walls, fences, poles, persons, and vehicles. 

\vspace{-3mm}
\subsubsection{Evaluation Metrics.} To evaluate patch performance against the MDE model, we introduce the relative error distance ($Rel.~Ed$) and region-affected error distance ($Ra_{Ed}$) metrics:
\begin{equation}
    Rel.~Ed = \frac{\sum (|d - d'| / d) \odot m_{patch}}{\sum m_{patch}},
\label{eq:rel_ed_metrics}
\end{equation}
\begin{equation}
    Ra_{Ed} = \frac{\sum B((|d - d'| / d) > Rel.~Ed_{thres}) \odot m_{road}}{\sum m_{patch}}.
\label{eq:ra_ed_metrics}
\end{equation}
Here, $d$ and $d'$ are the distance predictions for clean and adversarial images, $m_{patch}$ is the patch mask, $m_{road}$ is the road mask, $B()$ returns 1 if the condition is met, and $Rel.~Ed_{thres}$ is the threshold for the affected area set to 0.25. $Rel.~Ed$ measures the mean error caused by the adversarial patch, while $Ra_{Ed}$ gauges the ratio of affected road regions relative to the mask area. These metrics refine $E_d$ and $R_a$ from prior MDE patch attacks \cite{cheng_physical_2022,guesmi_aparate_2023,guesmi_saam_2024}.

To evaluate patch performance against the SS model, we use the Attack Success Rate ($ASR$) for untargeted ($ASR_{ua}$) and targeted ($ASR_{ta}$) attacks:
\begin{equation}
    ASR = 
    \begin{cases} 
      ASR_{ua} = \frac{\sum B(s' \neq s_{road}) \odot m_{patch} \odot s_{road}}{ \sum m_{patch} \odot s_{road}}, & \text{if untargeted} \\
      ASR_{ta} = \frac{\sum B(s' = c \in C) \odot m_{patch} \odot s_{road}}{ \sum m_{patch} \odot s_{road}}, & \text{if targeted}
    \end{cases}
\label{eq:asr_combined}
\end{equation}
Here, $s'$ is the semantic prediction of the adversarial image, $s_{road}$ is the road prediction of the original image, and $c$ represents target labels for targeted attacks. Multiplying $m_{patch}$ by $s_{road}$ ensures evaluation only on valid road predictions within patch areas.

Finally, we adapt the $Ra$ metric for the SS model to measure the ratio of affected error regions in the road area relative to the mask area:
\begin{equation}
    Ra_{ss} = 
    \begin{cases} 
      Ra_{ua} = \frac{\sum B(s' \neq s_{road}) \odot s_{road}}{\sum m_{patch}}, & \text{if untargeted} \\
      Ra_{ta} = \frac{\sum B(s' = c \in C) \odot s_{road}}{\sum m_{patch}}. & \text{if targeted}
    \end{cases}
\label{eq:ra_ss_combined}
\end{equation}
Unlike $ASR$, $Ra$ also considers the error outside the patch area.

\subsection{Experiment Results}

\subsubsection{Patch Effectiveness.} 
\begin{figure*}[!t]
\begin{center}
\vspace{-5mm}
\centerline{\includegraphics[width=1.025\textwidth]{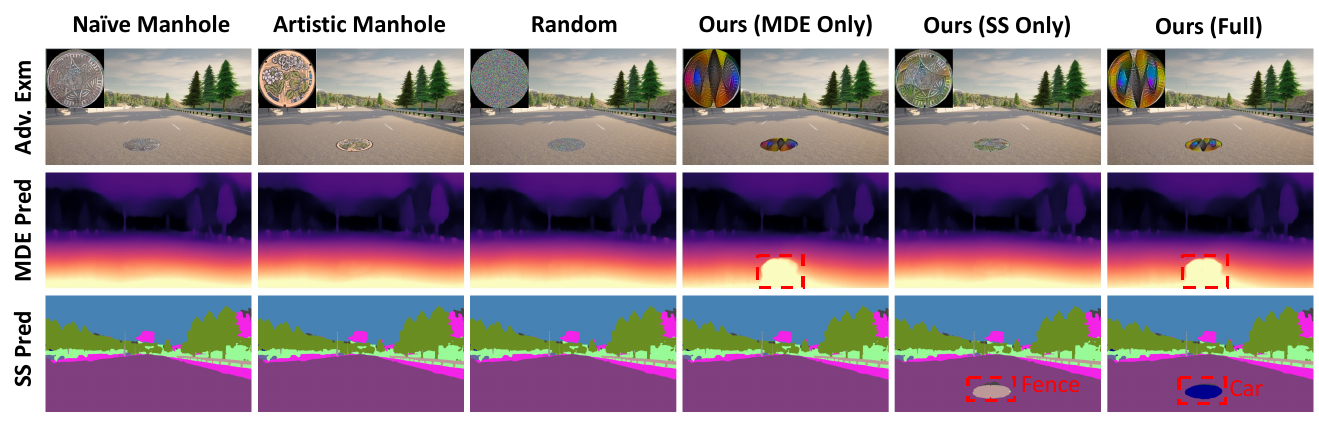}}
\caption{Qualitative results of our patch effectiveness: The results illustrate that naive, artistic, and random manholes do not affect the model predictions, while our methods successfully fool the models. Zoom for details.}
\label{fig:effectiveness_results}
\vspace{-1cm}
\end{center}
\end{figure*}

We evaluate our generated adversarial manhole on test sets, comparing it to baselines like naive, artistic, and random pattern manholes. We also compare it to related work, such as Adversarial Patch \cite{yamanaka_adversarial_2020} and SemSegAdvPatch \cite{nesti_evaluating_2022}, using their official code but with our settings.
As shown in Tab. \ref{tab:effectiveness_results} and Fig. \ref{fig:effectiveness_results}, the models are robust against the baselines, correctly predicting depth and segmentation as normal roads. While Adversarial Patch and SemSegAdvPatch disturb MDE and SS models, their performance declines due to uncovered manhole orientation during optimization. Note that SemSegAdvPatch supports only untargeted attacks, and we used the EOT variant since the manhole location isn’t fixed. Our adversarial manholes, however, can effectively fool the models based on the applied adversarial loss.
If $L_d$ is applied, the adversarial manhole causes an average relative error distance of $43\%$ to the MDE model. When $L_{ua}$ and $L_{ta}$ are applied, it achieves a $96\%$ attack success rate in fooling segmentation. Combining all adversarial losses, the manhole effectively attacks both MDE and SS with similar impact. Notably, the adversarial manhole can cause an error to MDE larger than the patch size, affecting a region $239\%$ relative to the patch size, despite the loss focusing only on the patch area.

\begin{table}[!t]
\vspace{-5mm}
\caption{Evaluation results of patches effectiveness in our test sets.}
\label{tab:effectiveness_results}
\begin{center}
\begin{tabular}{|l|c|c|c|c|c|c|}
\hline
\multirow{2}{*}{Evaluated Patches} & \multicolumn{2}{c|}{MonoDepth2 \cite{monodepth2} (MDE)} & \multicolumn{4}{c|}{DDRNet \cite{ddrnet} (SS)}                               \\ 
\cline{2-7}                        & $~~Rel.~Ed~~$       & $Ra_{Ed}$       & $ASR_{ua}$      & $ASR_{ta}$      & ~$Ra_{ua}$~  & ~$Ra_{ta}$~  \\ \hline 
Naïve Manhole                      & 0.03              & 0.02              & 0.00          & 0.00          & 0.00          & 0.00          \\
Artistic Manhole                   & 0.04              & 0.04              & 0.00          & 0.00          & 0.00          & 0.00          \\
Random Patch                       & 0.03              & 0.01              & 0.00          & 0.00          & 0.00          & 0.00          \\ \hline
Adversarial Patch \cite{yamanaka_adversarial_2020}  & 0.35     &  1.62     & 0.00          & 0.00          & 0.00          & 0.00          \\
Ours MDE ($L_{d}$+$L_{tv}$+$L_{c}$) & \textbf{0.43}     & \textbf{2.39}     & 0.00          & 0.00         & 0.00          & 0.00          \\ \hline
SemSegAdvPatch \cite{nesti_evaluating_2022}  & 0.06     &  0.04     & 0.40          & 0.04          & 0.95          & 0.09          \\
Ours SS ($L_{ua}$+$L_{ta}$+$L_{tv}$+$L_{c}$)  & 0.05              & 0.05              & \textbf{0.96} & 0.95          & \textbf{1.08} & 0.96 \\ \hline
\textbf{Ours All ($L_{total}$)}        & \textbf{0.43}     & 2.34              & \textbf{0.96} & \textbf{0.96} & 1.02          & \textbf{0.98}  \\ \hline
\end{tabular}
\end{center}
\end{table}

\vspace{-3mm}
\subsubsection{Patch Robustness.} 
\begin{figure}[!t]
    \vspace{-3mm}
    \centering
    \begin{subfigure}{0.24\textwidth}
        \includegraphics[width=\linewidth]{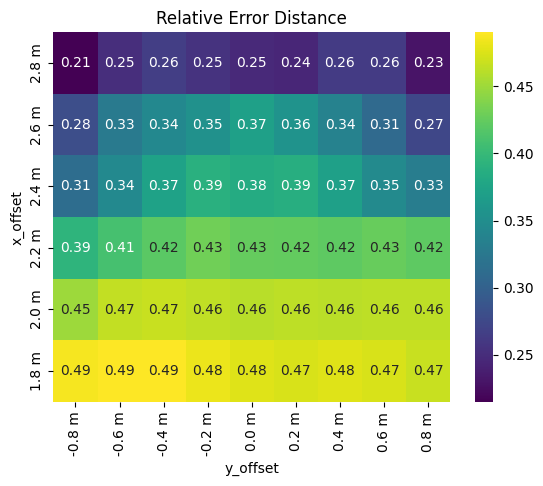}
        \caption{$Rel.~Ed$}
    \end{subfigure}
    \hfill
     \begin{subfigure}{0.24\textwidth}
        \includegraphics[width=\linewidth]{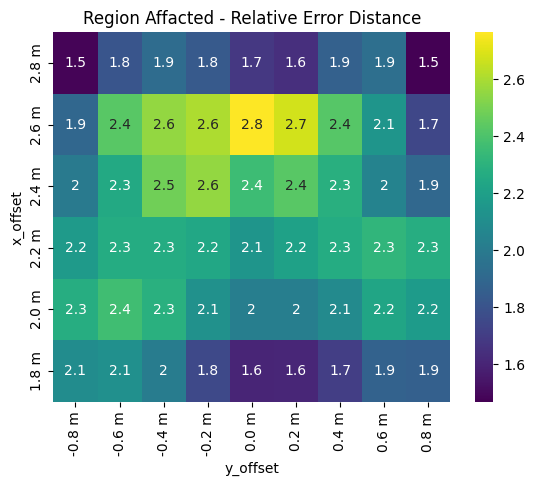}
        \caption{$Ra_{Ed}$}
    \end{subfigure}
    \hfill
    \begin{subfigure}{0.24\textwidth}
        \includegraphics[width=\linewidth]{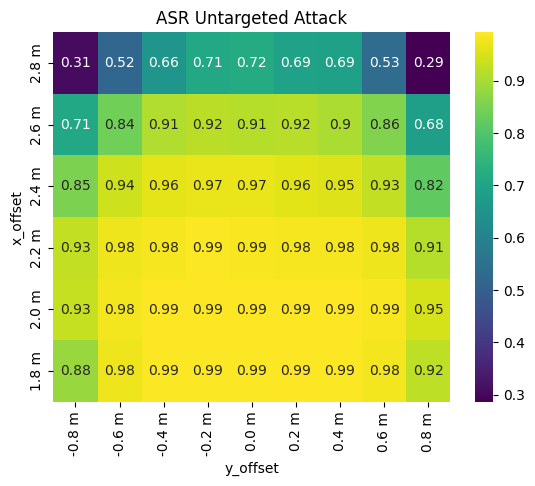}
        \caption{$ASR_{ua}$}
    \end{subfigure}
    \hfill
    \begin{subfigure}{0.24\textwidth}
        \includegraphics[width=\linewidth]{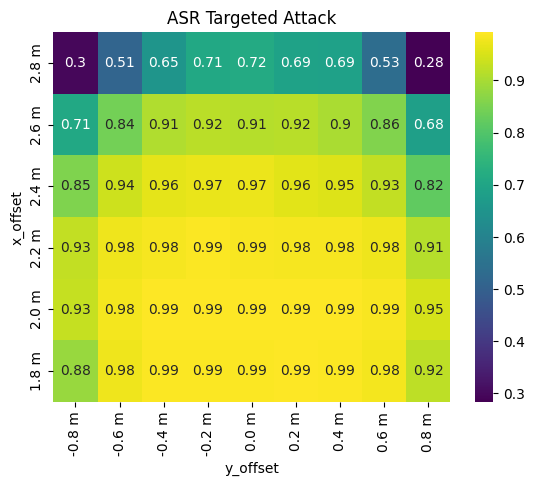}
        \caption{$ASR_{ta}$}
    \end{subfigure}
    \hfill
    \caption{Robustness evaluation against different placements. Zoom for details.}
    \label{fig:robustness_against_placements}
\end{figure}

To evaluate the robustness of our adversarial manhole against different placements on the road surface, we conducted experiments across various offsets. 
As summarized in Fig. \ref{fig:robustness_against_placements}, our patch remains robust regardless of location but performs best when centrally located and near the camera.
For MDE metrics, the relative error distance decreases linearly as the patch is placed further away, from 48\% relative error to 25\%. However, the region-affected error for MDE shows optimal performance when the patch is 2.8m from the camera's center. We observe this because the affected error region extends from the patch region to the bottom of the image. In contrast, the patch demonstrates a more robust performance on the SS model regardless of position, with an average attack success rate of 98\% except for edge placements.

\begin{figure*}[!t]
\begin{center}
\centerline{\includegraphics[width=1.025\textwidth]{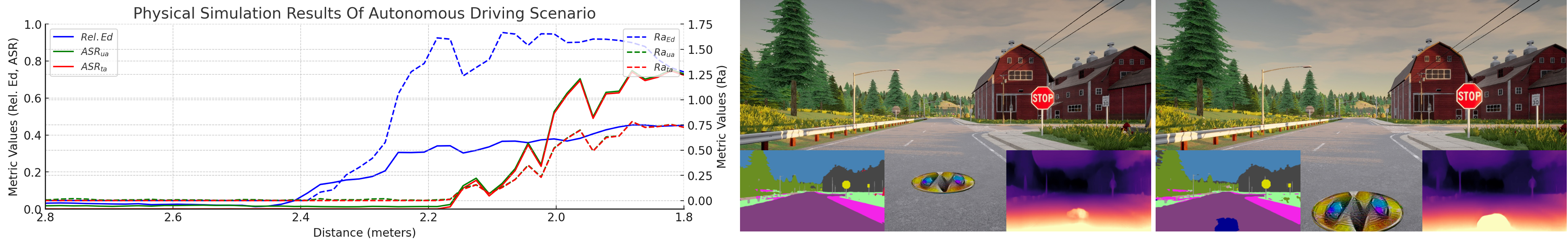}}
\caption{Physical simulation of approaching car in CARLA. Zoom for details.}
\label{fig:physical_simulation_results}
\vspace{-1cm}
\end{center}
\end{figure*}

\vspace{-3mm}

\subsubsection{Physical Simulation.}
We perform physical evaluation by applying the adversarial patch as a manhole cover in the CARLA simulator. We capture video of an approaching car to mimic an autonomous driving scenario. As illustrated in Fig. \ref{fig:physical_simulation_results}, the adversarial manhole begins affecting the MDE model from a distance of 2.4 meters and the SS model from 2.2 meters. The attack’s effectiveness increases as the camera approaches the manhole, reaching a peak performance of 45\% relative error distance and 166\% affected region for MDE, and a 75\% attack success rate with a 78\% affected region for SS. These results highlight the substantial impact of the adversarial patch in a more realistic setting.

\begin{table}[!b]
\centering
\vspace{-5mm}
\caption{Evaluation results of targeting different models.}
\label{tab:different_models}
\begin{tabular}{|l|l|c|c|c|c|c|c|}
\hline
\multirow{2}{*}{Target MDE Model} & \multirow{2}{*}{Target SS Model} & \multicolumn{2}{c|}{MDE Metrics} & \multicolumn{4}{c|}{SS Metrics} \\
\cline{3-8}
& & $Rel.~Ed$ &$ Ra_{Ed}$ & $ASR_{ua}$ & $ASR_{ta}$ &$ Ra_{ua}$ &$ Ra_{ta}$ \\
\hline
MonoDepth2 \cite{monodepth2} & DDRNet \cite{ddrnet} & \textbf{0.43} & \textbf{2.34} & 0.96 & 0.96 & 1.02 & 0.98 \\
DepthHint \cite{depthhints} & DDRNet \cite{ddrnet} & 0.33 & 0.83 & 0.97 & 0.97 & 1.01 & 0.98 \\
ManyDepth \cite{manydepth} & DDRNet \cite{ddrnet} & 0.28 & 1.51 & 0.96 & 0.96 & 1.02 & 0.97 \\
\hline
MonoDepth2 \cite{monodepth2} & PSPNet \cite{zhao_pyramid_2017} & 0.40 & 2.03 & \underline{0.94} & \underline{0.93} & 1.00 & \underline{0.95} \\
DepthHint \cite{depthhints}  & PSPNet \cite{zhao_pyramid_2017} & \underline{0.24} & \underline{0.53} & \underline{0.94} & 0.94 & \underline{0.98} & \underline{0.95} \\
ManyDepth \cite{manydepth}  & PSPNet \cite{zhao_pyramid_2017} & 0.25 & 1.32 & \underline{0.94} & \underline{0.93} & 1.00 & \underline{0.95} \\
\hline
MonoDepth2 \cite{monodepth2} & ICNet \cite{ferrari_icnet_2018} & 0.42 & 2.32 & \textbf{0.99} & \textbf{0.99} & 2.65 & 2.59 \\
DepthHint \cite{depthhints}  & ICNet \cite{ferrari_icnet_2018} & 0.31 & 0.76 & \textbf{0.99} & \textbf{0.99} & 2.98 & 2.92 \\
ManyDepth \cite{manydepth}  & ICNet \cite{ferrari_icnet_2018} & 0.29 & 1.67 & \textbf{0.99} & \textbf{0.99} & \textbf{3.13} & \textbf{3.09} \\
\hline
\end{tabular}
\vspace{-5mm}
\end{table}

\vspace{-3mm}
\subsubsection{Applicability to Different Models.}
We regenerated adversarial manholes with different target models to assess their applicability against various MDE and SS models. As summarized in Table \ref{tab:different_models}, our patch has the highest overall impact when targeting MonoDepth2 with DDRNet for MDE metrics (43\% relative error distance with $2\times$ affected regions) and ManyDepth with ICNet for SS metrics (99\% attack success rate with $3\times$ affected regions). However, the lowest overall impact is observed when targeting DepthHint and PSPNet (24\% relative error distance with 53\% affected regions for MDE and 94\% attack success rate with 95\% affected regions for SS), indicating that these models are more robust.

\vspace{-3mm}
\subsubsection{Ablation Studies.} 
We performed ablation studies by removing each proposed module, regenerating the patch, and applying random brightness and contrast adjustments to test robustness. As shown in Table \ref{tab:abs_studies}, removing any module decreased performance, except for content loss ($L_{c}$), which trades off stealthiness and effectiveness. Optimizing the patch at a fixed offset ($t_O$) reduced robustness against varied placements. Replacing DPM with random spatial and perspective transformations (*no $DPM$) from previous works \cite{yamanaka_adversarial_2020,nesti_evaluating_2022,cheng_physical_2022,guesmi_aparate_2023,guesmi_saam_2024} significantly impaired performance in the manhole setting. These findings emphasize the need to cover expected transformations during optimization, as noted in EOT \cite{athalye2018synthesizing}.

\begin{table}[!t]
\centering
\caption{Ablation studies with the full baseline under random distortion during evaluation. Values in parentheses indicate the difference from the full baseline.}
\label{tab:abs_studies}
\resizebox{\columnwidth}{!}{
\begin{tabular}{|l|c|c|c|c|c|c|}
\hline
\multirow{2}{*}{Modules} & \multicolumn{2}{c|}{MonoDepth2 \cite{monodepth2} (MDE)} & \multicolumn{4}{c|}{DDRNet \cite{ddrnet} (SS)} \\
\cline{2-7}
 & $Rel.~Ed$ & $Ra_{Ed}$ & $ASR_{ua}$ & $ASR_{ta}$ & $Ra_{ua}$ & $Ra_{ta}$ \\
\hline
full & 0.42 & 2.18 & 0.97 & 0.96 & 1.01 & 0.98 \\
fixed $t_O$ & \underline{0.19} (\underline{-0.24}) & \underline{0.51} (\underline{-1.68}) & 0.27 (-0.69) & 0.26 (-0.71) & 0.24 (-0.78) & 0.20 (-0.78) \\
*no $DPM$ & 0.25 (-0.17) & 0.93 (-1.25) & \underline{0.01} (\underline{-0.95}) & \underline{0.00} (\underline{-0.96}) & \underline{0.02} (\underline{-0.99}) & \underline{0.00} (\underline{-0.98}) \\
no $A_t$ & 0.26 (-0.16) & 1.12 (-1.06) & 0.41 (-0.55) & 0.40 (-0.56) & 0.45 (-0.56) & 0.42 (-0.56) \\
no $A_f$ & 0.42 (0.00) & 2.13 (-0.05) & 0.90 (-0.07) & 0.90 (-0.07) & 0.96 (-0.05) & 0.93 (-0.05) \\
no $L_{tv}$ & 0.42 (0.00) & 2.13 (-0.06) & 0.91 (-0.05) & 0.91 (-0.05) & 0.94 (-0.07) & 0.91 (-0.07) \\
no $L_c$ & \textbf{0.49 (0.06)} & \textbf{2.72 (0.53)} & \textbf{0.98 (0.02)} & \textbf{0.98 (0.02)} & \textbf{1.07 (0.06)} & \textbf{1.02 (0.04)} \\
\hline
\end{tabular}
}
\vspace{-3mm}
\end{table}

\vspace{-1mm}
\section{Conclusion}
\vspace{-3mm}
We presented a novel adversarial attack using practical patches that mimic manhole covers to deceive MDE and SS models in autonomous driving systems. Our patches achieved a 43\% relative error in MDE and a 96\% attack success rate in SS. Physical simulations validated the effectiveness of our adversarial patches in an autonomous driving scenario. Ablation studies confirmed the critical role of our proposed modules and highlighted the applicability of patches in different models. 
Our work underscores the vulnerability and emphasizes the need for robust defense mechanisms, 
such as certified defense against patch attacks \cite{yatsura2023certified}.

\vspace{-2mm}

{
\bibliographystyle{splncs04}
\bibliography{references}
}

\end{document}